\documentclass[runningheads]{llncs}
\usepackage[T1]{fontenc}
\usepackage{graphicx}
\usepackage[misc]{ifsym}

\usepackage{multirow}
\usepackage{rotating}
\usepackage{verbatim} 
\usepackage{booktabs}

\begin{document}
\title{First Experiences with the Identification of People at Risk for Diabetes in Argentina using Machine Learning Techniques}
%
%\titlerunning{}
\author{Enzo Rucci\inst{1,2}\orcidID{0000-0001-6736-7358} \Letter
\and Gonzalo Tittarelli \inst{3}
   \and Franco Ronchetti\inst{1,2}\orcidID{0000-0003-3173-1327}
Jorge F. Elgart\inst{4}\orcidID{0000-0002-6101-1219} \and 
Laura Lanzarini\inst{1}\orcidID{0000-0001-7027-7564} \and
Juan José Gagliardino\inst{4}\orcidID{0000-0001-7332-6536}
}
\institute{  III-LIDI, Facultad de Informática, UNLP – CIC. La Plata (1900), Argentina \\
\email{\{erucci,fronchetti,laural\}@lidi.info.unlp.edu.ar} \\
\and Comisión de Investigaciones Científicas (CIC). La Plata (1900), Argentina \\
\and Facultad de Informática, UNLP. La Plata (1900), Bs As, Argentina \\
\and CENEXA, Facultad de C. Médicas, UNLP-CONICET. La Plata (1900), Argentina \\
\email{\{jelgart,jjgagliardino\}@cenexa.org}
\and CONICET. La Plata (1900), Argentina \\
}
\maketitle              % typeset the header of the contribution
\begin{abstract}
Detecting Type 2 Diabetes (T2D) and Prediabetes (PD) is a real challenge for medicine due to the absence of pathogenic symptoms and the lack of known associated risk factors. Even though some proposals for machine learning models enable the identification of people at risk, the nature of the condition makes it so that a model suitable for one population may not necessarily be suitable for another. In this article, the development and assessment of predictive models to identify people at risk for T2D and PD specifically in Argentina are discussed. First, the database was thoroughly preprocessed and three specific datasets were generated considering a compromise between the number of records and the amount of available variables. After applying 5 different classification models, the results obtained show that a very good performance was observed for two datasets with some of these models. In particular, RF, DT, and ANN demonstrated great classification power, with good values for the metrics under consideration. Given the lack of this type of tool in Argentina, this work represents the first step towards the development of more sophisticated models.

\end{abstract}

\begin{center}
\texttt{This version of the contribution has been accepted for publication, after peer review (when applicable) but is not the Version of Record and does not reflect post-acceptance improvements, or any corrections. The
Version of Record is available online at: \url{https://doi.org/10.1007/978-3-031-62245-8\_16}. Use of this
Accepted Version is subject to the publisher’s Accepted Manuscript terms of use
\url{https://www.springernature.com/gp/open-research/policies/accepted-manuscript-terms}}
\end{center}

\clearpage

\section{Introduction}
\label{sec:intro}

%La Diabetes Tipo 2 (DT2) es una enfermedad crónica caracterizada por niveles elevados de glucemia que se manifiesta cuando el páncreas endocrino es incapaz de producir la cantidad de insulina suficiente que requieren sus tejidos. Debido a su creciente prevalencia en combinación con su elevado costo de atención, constituye un serio problema de salud pública, por lo que se han realizado grandes esfuerzos por desarrollar estrategias efectivas para su prevención y tratamiento oportuno, así como para evitar el desarrollo de sus complicaciones crónicas.

Type 2 Diabetes (T2D) is a chronic disease characterized by high blood glucose levels; it manifests when the endocrine pancreas is unable to produce enough insulin for body tissues~\cite{ManualDM}. Due to its increasing prevalence in combination with its high cost of care~\cite{Bolin2009,Williams2002}, it represents a serious public health problem, and great efforts have been made to develop effective strategies for its prevention and timely treatment, as well as to avoid the development of its chronic complications.

%Resulta importante reconocer que las consecuencias negativas de esta enfermedad comienzan en una etapa previa conocida como prediabetes (PDM), la cual está definida como una elevación en la concentración de glucosa en sangre más allá de los niveles normales pero sin alcanzar los valores diagnósticos de diabetes. La PDM se manifiesta a través de la Glucemia en Ayunas Alterada (GAA), la Tolerancia a la Glucemia Alterada (TGA) y la combinación de ambas~\cite{ADA-S3}. La PDM implica un riesgo elevado de desarrollar DT2 del orden del 30\%~\cite{DPP2002} y de 70\%~\cite{Eddy2005} en los siguientes 4 y 30 años, respectivamente. 

It is worth noting that the negative consequences of this disease start in a previous stage known as prediabetes (PD), which is characterized by elevated blood glucose concentration higher than normal levels, but without reaching the diagnostic values for diabetes. PD involves Impaired Fasting Glucose (IFG), Impaired Glucose Tolerance (IGT), and a combination of both~\cite{ADA-S3}. PD has a high risk of progressing into T2D — around 30\%~\cite{DPP2002} and 70\%~\cite{Eddy2005} in the following 4 and 30 years, respectively.

%El desarrollo de la DT2 es un proceso lento y progresivo condicionado por factores genéticos, ambientales y de comportamiento. Actualmente no existe una cura definitiva para esta enfermedad. Sin embargo, varios estudios han demostrado que se puede prevenir o demorar su aparición en personas con PDM a través de la adopción de un estilo de vida saludable (plan de alimentación y práctica regular de actividad física) y/o asociado con la ingesta de diversos fármacos~\cite{DPP2002,Tuomilehto2001}. En Argentina, una iniciativa de este tipo es el Programa Piloto para la Prevención Primaria de Diabetes en la provincia de Buenos Aires (PPDBA) desarrollado por el CENEXA (UNLP-CONICET) y financiado por el Ministerio de Ciencia y Tecnología de la Nación, la empresa SANOFI y el CONICET~\cite{PPDBA2016}.

The development of T2D is a slow and progressive process conditioned by genetic, environmental, and behavioral factors. Currently, there is no definitive cure for this disease. However, several studies have shown that its onset can be prevented or delayed in people with PD through the adoption of a healthy lifestyle (eating plan and regular physical activity) and/or associated with the intake of various drugs~\cite{DPP2002,Tuomilehto2001}. In Argentina, an initiative of this type is the Pilot Program for the Primary Prevention of Diabetes in the province of Buenos Aires (PPDBA), developed by CENEXA (UNLP-CONICET) and financed by the Ministry of Science and Technology of the Nation, the company SANOFI and CONICET~\cite{PPDBA2016}.

%La detección de DT2 y PDM representa un verdadero desafío para la medicina debido a la ausencia de síntomas patogenómicos y a la falta de conocimiento de los factores de riesgo asociados. Es por eso que frecuentemente una persona pueda pasar meses (o incluso años) sin saber que se encuentra en riesgo. En ese sentido, estadísticas publicadas en el año 2018 por la Federación Internacional de Diabetes muestran que aproximadamente un 50\% de la población mundial desconoce su enfermedad~\cite{FIDAtlas2018}. Esto explica la necesidad de contar con un método de detección simple y preciso. En consecuencia, este artículo propone desarrollar y evaluar modelos predictivos basados en aprendizaje automático (AA) que permitan identificar personas con riesgo de diabetes y PDM en la población argentina considerando como base de datos la correspondiente al PPDBA. %REVISAR ESTA PARTE. En particular, se considerarán las técnicas Reglas de Clasificación y Naive-Bayes.

Detecting T2D and PD is a real challenge for medicine due to the absence of pathogenic symptoms and the lack of known associated risk factors. This is why individuals may often go months (or even years) without knowing that they are at risk. In this sense, statistics published in 2018 by the International Diabetes Federation show that approximately 50\% of the world's population is unaware of their disease~\cite{FIDAtlas2018}. This explains the need for a simple and accurate detection method. Thus, this article proposes the development and assessment of predictive models based on machine learning (ML) that will allow identifying people at risk of T2D and PD in Argentina. To this end, the PPDBA database will be used.  

This paper is an extended and thoroughly revised version of~\cite{RucciCACIC2023}. The work has been extended by providing: 
(1) the proposal of a new segmentation that only considers clinical data from the PPDBA database and the training of several models that use it as input; and (2) a performance analysis of the new models and a cost-benefit analysis with the previous ones.

%El resto del artículo se organiza de la siguiente forma. La Sección~\ref{sec:back} introduce el marco referencial para este trabajo. Luego, la Sección~\ref{sec:imple} describe el procesamiento realizado a la base de datos mientras que la Sección~\ref{sec:results} describe y analiza los resultados obtenidos. Finalmente, la Sección~\ref{sec:conc} presenta las conclusiones junto al trabajo futuro.

The remaining sections of this article are organized as follows: In Section~\ref{sec:back}, the background for this work is presented. In Section~\ref{sec:imple}, database processing is described, while in Section~\ref{sec:results}  the results obtained are analyzed. Finally, in Section~\ref{sec:conc}, conclusions and possible future work are presented.

\section{Background}
\label{sec:back}

\subsection{Diabetes and Prediabetes — Risk Factors and Diagnosis}
\label{sec:dm}

%El desarrollo de DT2 es un proceso lento y progresivo que se encuentra condicionado por factores genéticos, ambientales y de comportamiento. Entre los factores de riesgo se encuentran, entre otros, el género, el índice de masa corporal (IMC), la circunferencia de  cintura, los hábitos de alimentación, la práctica de actividad física, la edad, los antecedentes familiares de diabetes (incluida gestacional), la etnia y los trastornos del sueño. Por su parte, la PDM representa un estado previo a la DT2 y su progresión puede prevenirse e incluso revertirse mediante la adopción de estilos de vida saludables~\cite{Vistisen2019}. Si además consideramos que la PDM no es una pre-enfermedad, pues ya presenta disfunciones metabólicas, resulta sumamente importante identificar personas con PDM tanto como lo es poder hacerlo con las que ya tienen DT2 no diagnosticada. 

%El diagnóstico se realiza por medio de análisis de sangre y  cualquier persona con presencia de los síntomas o factores de riesgo asociados debe ser examinada. En Argentina, habitualmente se emplea la Prueba de Tolerancia Oral a la Glucosa (PTOG) para determinar si una persona posee DT2, PDM o ninguna de ellas~\footnote{Esta prueba comienza con una extracción de sangre de la persona en ayunas. Luego, se le pedirá que tome un líquido que contiene una cierta cantidad de glucosa. A  continuación, se le tomarán muestras de sangre nuevamente cada 30 a 60 minutos después de ingerir la solución. El examen es costoso económicamente y puede demorar hasta 3 horas.}.

 The development of T2D is a slow and progressive process that is conditioned by genetic, environmental, and behavioral factors. Risk factors include gender, body mass index (BMI), waist circumference, eating habits, physical activity, age, family history of diabetes (including gestational), ethnicity, sleep disorders, and so forth. PD corresponds to a state before T2D, and its progression can be prevented and even reversed by adopting healthy lifestyles~\cite{Vistisen2019}. If we also consider that PD is not a pre-disease state, since it already presents metabolic dysfunctions, it is extremely important to identify people with PD as much as it is to do so with those who already have undiagnosed T2D.

Diagnosis is based on blood tests, and anyone presenting the associated symptoms or risk factors should be examined. In Argentina, the Oral Glucose Tolerance Test (OGTT) is usually used to determine if a person has T2D, PD, or neither~\footnote{For this test, a blood sample is drawn on a fasting state. Then, the subject is asked to drink a liquid containing a certain amount of glucose. After this, another blood sample is taken every 30 to 60 minutes after intake. The exam is financially expensive and can take up to 3 hours.}.

%Como estas características pueden variar de una población a otra, un modelo predictivo que resulte adecuado para una población, no necesariamente lo será para otra.

\subsection{Related Works}
\label{sec:relwork}

In the last decade, numerous models have been proposed to identify undiagnosed diabetes and/or PD using ML techniques. These are based on various clinical and laboratory variables linked to relevant risk factors. Most of them \cite{Sarojni2010} \cite{AlJarullah2011} \cite{Hashi2017} \cite{Jayanthi2017} \cite{Maniruzzaman2017} \cite{Mercaldo2017} \cite{Jayanthi2017} \cite{Yuvaraj2017} \cite{Dey2018} \cite{Wei2018} \cite{Mir2018} \cite{Sisodia2018} \cite{Kaur2018} \cite{Zou2018} \cite{Wu2018} \cite{Sneha2019} use a diabetes database known as PIMA Indian Diabetes (PID) from the repository of the University of California Irvine~\footnote{\url{https://archive.ics.uci.edu/ml/index.php}}, USA. This dataset contains records of women from the Pima indigenous people (USA) and is made up of 8 attributes related to risk factors for developing the disease. It has a total of 786 records; however, this number is reduced by half if entries with null values are removed. These works are more of a proof of concept than a real implementation and, in general, have been aimed at improving the performance (accuracy) of diabetes classification algorithms.

\begin{comment}
Son pocos los trabajos que no emplean la base de datos PID. Entre ellos se encuentran ~\cite{Heikes2008} \cite{Yu2010} \cite{Dinh2019} quienes propusieron modelos para identificar diabetes y prediabetes usando datos de la Encuesta NHANES de EEUU. %Nacional de Examen de Salud y Nutrición de EEUU (NHANES).
Recientemente, \cite{Xie2019} también estudió la aplicación de distintas técnicas de AA para la detección de diabetes no diagnosticada en la población estadounidense aunque empleando la base de datos %del Sistema de Vigilancia de Factores de Riesgo del Comportamiento(BRFSS).
BRFSS. En forma similar, \cite{Meng2013} presentó diferentes modelos predictivos de estas enfermedades para la población china usando una base de datos \textit{ad-hoc}. A diferencia de los anteriores, \cite{Choi2014} hizo hincapié en el desarrollo y validación de modelos predictivos únicamente para prediabetes. %Para ello, se valió de los datos de la la Encuesta Nacional de Examen de Salud y Nutrición de Corea (KNHANES).
\end{comment}

Only a few of these works do not use the PID database. These include ~\cite{Heikes2008} \cite{Yu2010} \cite{Dinh2019} , whose authors proposed models to identify diabetes and prediabetes using data from the US NHANES Survey. Recently, the authors in  \cite{Xie2019} also studied the application of different ML techniques for detecting undiagnosed diabetes in the US population using the BRFSS database. Similarly, the authors in  \cite{Meng2013} presented various predictive models for these diseases for the Chinese population using an ad-hoc database. In  \cite{Choi2014}, on the other hand, the development and validation of predictive models focused only on prediabetes.

%Como se explicó en la Sección~\ref{sec:dm}, el desarrollo de DT2  se encuentra condicionado por factores que pueden variar de una población a otra. Es por lo que un modelo predictivo que resulte adecuado para una población, no necesariamente lo será para otra. Este estudio representa el primer paso hacia modelos predictivos específicos para la población argentina.

As explained in Section~\ref{sec:dm}, the onset of T2D is conditioned by factors that may vary from one population to another. This is why a predictive model that is suitable for one population will not necessarily be suitable for another. This study is the first step toward building specific predictive models for the Argentine population.

\section{Implementation}
\label{sec:imple}

\subsection{Dataset}

\subsubsection{Description}

%Los modelos predictivos serán desarrollados a partir de la base de datos del programa PPDBA~\cite{PPDBA2016}, la cual cuenta con 1316 registros de personas.
%Cada registro corresponde a una persona que mediante PTOG fue identificada como diabética, prediabética o sin ninguna de ellas. AdeMore than datos de laboratorio (hemoglobina glicosilada; colesterol total, HDL y LDL;  triglicéridos y creatinina), se cuenta con variables clínicas asociadas a los factores de riesgo de estas enfermedades tales como el sexo; la edad; el Indice de Masa Corporal (IMC); la presión arterial; los antecedentes familiares de diabetes; los hábitos alimenticios y de actividad física; entre otros. 

Predictive models will be developed using the PPDBA program database~\cite{PPDBA2016}, which has 1316 records. Each record corresponds to a person identified as diabetic, prediabetic or disease-free through OGTT. In addition to laboratory data (glycated hemoglobin, total cholesterol, HDL cholesterol, LDL cholesterol, triglycerides, and creatinine), there are clinical variables associated with the risk factors for these diseases such as sex, age, Body Mass Index (BMI), blood pressure, family history of diabetes, eating habits, physical activity, and so forth.

\subsubsection{Characterization}

\begin{comment}

De los 1316 registros actuales, 80 debieron ser descartados ya que omitían valores requeridos de glucemia para poder calcular el resultado de la PTOG (no es posible determinar la \texttt{clase}). 
La Tabla~\ref{tab:my-table2} presenta una breve descripción estadística de los 1236 registros disponibles. Se puede notar que hay varias variables  que presentan nulos, los cuales se analizan con mayor profundidad en la sección siguiente. También se puede observar que: hay más personas del sexo femenino que del masculino; la mayoría de las personas tienen entre 45-64 años; la mayoría de las personas tienen un IMC mayor a 30 kg/m$^2$; la mayoría de las personas tienen una circunferencia de cintura de More than 102cm y de 88cm para el sexo masculino y femenino, respectivamente; la mayoría de las personas realizan actividad física; consumen vegetales, frutas y hortalizas; no toman medicación para controlar hipertensión; sí le encontraron hiperglucemia en algún control; y alguno de sus familiares (de primer o segundo grado) tiene diabetes; las variables asociadas a glucemia basal, colesterol HDL, triglicéridos y creatinina basal parecieran tener una amplia dispersión; el resto no; en cuanto a la clase, la mitad de las personas no están en riesgo de tener prediabetes o diabetes.
\end{comment}

Out of the 1316 current records, 80 had to be discarded due to the lack of blood glucose data to calculate the OGTT (\texttt{class} could not be determined). Table~\ref{tab:my-table2} presents a brief statistical description of the 1236 remaining records. There are several null variables, which are analyzed in greater depth in the following section. As it can be noted, there are more women than men, most subjects are between 45-64 years old; most with a BMI greater than 30 kg/m$^2$; most with a waist circumference greater than 102 cm (men) and 88 cm (women); most do physical activity; most eat vegetables and fruits; most do not take medications to control hypertension; most with a positive result for hyperglycemia at least once; most with at least one relative (first or second degree) with diabetes. Variables associated with baseline glycemia, HDL cholesterol, triglycerides, and baseline creatinine seem to have a wide dispersion; and as regards class, half of the subjects are not at risk for PD or T2D.

%\begin{comment}

% \usepackage{booktabs}
% \usepackage{multirow}
% \usepackage{graphicx}
% \usepackage{lscape}
\begin{table}[tb!]
\centering
\caption{Statistical overview of the dataset}
\label{tab:my-table2}
\resizebox{\columnwidth}{!}{%
\begin{tabular}{p{.3\columnwidth}p{.15\columnwidth}p{.35\columnwidth}p{.2\textwidth}}
\toprule
\textbf{Variable}                  & \textbf{\# null}  & \textbf{Metric}                         & \textbf{Result} \\ 
\midrule
\multirow{2}{*}{sex}              & \multirow{2}{*}{0} & Male (\%)                          & 395 (32\%)                    \\
                                   &                       & Female (\%)                           & 841 (68\%)                    \\
\hline
age                               & 564                   & Mean+SD                                & 57.23$\pm$8.8428                  \\
\hline
\multirow{4}{*}{age\_range}       & \multirow{4}{*}{0} & Less than 45 years (\%)                   & 205 (17\%)                    \\
                                   &                       & 45-54 years (\%)                         & 435 (35\%)                    \\
                                   &                       & 54-64 years (\%)                         & 429 (34\%)                    \\
                                   &                       & More than 64 years (\%)                    & 167 (14\%)                    \\
\hline
bmi                                & 619                   & Mean+SD                                & 31.65$\pm$6.314                   \\
\hline
\multirow{3}{*}{bmi\_range}        & \multirow{3}{*}{0} & Less than 25kg/m2                        & 93 (8\%)                      \\
                                   &                       & 25-30 kg/m2                             & 319 (26\%)                    \\
                                   &                       & More than 30 kg/m2                       & 824 (66\%9                    \\
\hline
waist\_circumference            & 1181                    & Mean+SD                               & 101.3091$\pm$13.55                \\
\hline
\multirow{3}{*}{waist\_circumference\_range}     & \multirow{3}{*}{0} & Less than 94/80 cm (M/F) (\%) & 56 (4\%)   \\
                                   &                       & M:   94-102cm / F: 80-88cm (\%)         & 205 (17\%)                    \\
                                   &                       & M: More than 102/88cm (M/F) (\%) & 975 (79\%)                    \\
\hline
\multirow{2}{*}{physical\_activity} & \multirow{2}{*}{0} & Yes (\%)                                 & 915 (74\%)                    \\
                                   &                       & No (\%)                                 & 321 (26\%)                    \\
\hline
\multirow{2}{*}{eat\_fruit\_vegetables}        & \multirow{2}{*}{0} & Yes (\%)                                    & 821 (66\%) \\
                                   &                       & No (\%)                                 & 415 (34\%)                    \\
\hline
\multirow{2}{*}{ht\_drugs}           & \multirow{2}{*}{0} & Yes (\%)                                    & 497 (40\%) \\
                                   &                       & No (\%)                                 & 739 (60\%)                    \\
\hline
\multirow{2}{*}{hyperglycemia}                   & \multirow{2}{*}{0} & Yes (\%)                                    & 999 (81\%) \\
                                   &                       & No (\%)                                 & 237 (19\%)                    \\
\hline
\multirow{3}{*}{family\_diagnosis} & \multirow{3}{*}{3} & No (\%)                                    & 395 (32\%) \\
                                   &                       & First degree (\%)                       & 412 (33\%)                    \\
                                   &                       & Second degree (\%)                      & 426 (34\%)                    \\
\hline
baseline\_glycemia                    & 0                  & Mean+SD                               & 104.36$\pm$27.28                  \\
\hline
postprandial\_glycemia                & 55                  & Mean+SD                               & 119.59$\pm$42.51                  \\
\hline
total\_cholesterol                  & 705                   & Mean+SD                               & 198.28$\pm$41.1                   \\
\hline
ldl\_cholesterol                    & 715                   & Mean+SD                               & 119.79$\pm$36.82                  \\
\hline
hdl\_cholesterol                    & 706                   & Mean+SD                               & 49.82$\pm$14.40                   \\
\hline
triglycerides                      & 705                   & Mean+SD                               & 151.4$\pm$95.59                   \\
\hline
baseline\_creatinine                  & 619               & 
Mean+SD                               & 1.117$\pm$5.8                                  \\
\hline
glycated\_hemoglobine           & 635                   & Mean+SD                               & 5.61$\pm$0.43                     \\
\hline
\multirow{3}{*}{ogtt\_result}   & \multirow{3}{*}{0} & No risk (\%)                         & 620 (50\%)                    \\
                                   &                       & Prediabetes (\%)                        & 480 (38\%)                    \\
                                   &                       & Diabetes (\%)                           & 136 (12\%)                    \\ \cmidrule(l){1-4} 
\end{tabular}%
}
\end{table}

\subsubsection{Cleanup}

\paragraph{Noise}

%Para analizar la presencia de ruido de las variables, se utilizó el método de Tukey para identificar los intervalos de valores atípicos leves y extremos. En base a lo anterior, se detectaron valores atípicos leves en las variables \texttt{edad} (4), \texttt{imc} (10), \texttt{circ\_\_\_de\_cintura} (1), \texttt{glucemia\_basal} (27), \texttt{glucemia\_pprandial} (46), \texttt{colesterol\_ldl} (7), \texttt{colesterol\_total} (9), \texttt{colesterol\_hdl} (11), \texttt{triglicéridos} (21) y \texttt{creatinina\_basal} (8). Además, se encontraron valores atípicos extremos en \texttt{glucemia\_basal} (51), \texttt{glucemia\_pprandial} (6), \texttt{trigliceridos} (12) y \texttt{creat\_basal} (6). 

To analyze the presence of noise in the variables, Tukey's method was used to identify the ranges for outliers. They were detected in the following variables: \texttt{age} (4), \texttt{bmi} (10), \texttt{waist\_circumference} (1), \texttt{baseline\_glycemia} (27), \texttt{postprandial\_glycemia} (46), \texttt{ldl\_cholesterol} (7), \texttt{total\_cholesterol} (9), \texttt{hdl\_cholesterol} (11), \texttt{triglycerides} (21), and \texttt{baseline\_creatinine} (8). Additionally, \textit{far-out} outliers were found in \texttt{baseline\_glycemia} (51), \texttt{postprandial\_glycemia} (6), \texttt{triglycerides} (12), and \texttt{baseline\_creatinine} (6).

\paragraph{Missing values}

%En la Tabla~\ref{tab:my-table2} se puede observar que no hay valores nulos en las variables cualitativas, a excepción de \texttt{le\_diag\_\_\_familiar}, que presenta 3 registros con valores nulos. No ocurre lo mismo con las variables cuantitativas, donde el faltante de valores es mucho mayor. Afortunadamente, para algunas de esas variables cuantitativas, sí se cuenta con una variable cualitativa asociada que permite conocer el rango en que se encuentra ese valor faltante. Esto ocurre concretamente en los casos de \texttt{edad}, \texttt{imc} y \texttt{circ\_\_de\_cintura}.  El resto de las variables que presentan nulos son: \texttt{glucemia\_pprandial}, \texttt{colesterol\_total}, \texttt{colesterol\_ldl}, \texttt{colesterol\_hdl}, \texttt{trigliceridos}, \texttt{creat\_basal} y \texttt{hem\_\_\_glicosilada}. 

Table~\ref{tab:my-table2} shows that there are no missing values in qualitative variables, except for \texttt{family\_diagnosis}, with 3 records with null values. In the case of quantitative variables, the number of missing values is much greater. Fortunately, some of these quantitative variables have an associated qualitative variable that limits the range for that missing value. This occurs specifically in the cases of \texttt{age}, \texttt{bmi}, and \texttt{waist\_circumference}. The other variables that have null values are: \texttt{postprandial\_glycemia}, \texttt{total\_cholesterol}, \texttt{ldl\_cholesterol}, \texttt{hdl\_cholesterol}, \texttt{triglycerides}, \texttt{baseline\_creatinine}, and \texttt{glycosylated\_hemoglobin}.

%Para las variables con nulos hay 3 opciones:

%\begin{enumerate}
 %   \item Ignorar el registro, lo que lleva a disminuir el tamaño de la muestra. Esta opción puede ser conveniente si la cantidad de registros que tienen valores nulos en esta variable es pequeña. De otro modo, se reduciría significativamente el tamaño de la muestra, perdiendo el aporte de las otras características.
  %  \item Eliminar la variable, lo que lleva a tener una característica menos para entrenar nuestro modelo. A diferencia del caso anterior, esta opción puede ser conveniente si la variable cuenta con muchos valores nulos. De esta manera, se prioriza mantener el tamaño de la muestra.
   % \item Reemplazar los valores nulos con algún valor especial que tenga sentido considerando el dominio, como la media, la mediana o la moda. O usar alguna técnica específica que permite calcular el valor probable (ej. regresor).
%\end{enumerate}

\subsubsection{Transformations}

\paragraph{Treatment of outliers}

Medical experts were consulted about the occurrence of outliers in lab tests. It was concluded that, even if these values in the database are statistically atypical, they are within the possible range for their respective tests. There is only one exception — 2 values in \texttt{baseline\_creatinine} (85 and 118) that are quite likely upload errors. Therefore, those two specific values were replaced with null values.

%Se consultó con expertos del dominio médico sobre las ocurrencias de valores atípicos en las determinaciones de laboratorio. Se concluyó que, si bien son valores que estadísticamente pueden ser considerados atípicos, sí se encuentran dentro de los posibles valores extremos para estas determinaciones. La excepción son 2 valores en \texttt{creatinina\_basal} ('85', '118') que efectivamente  corresponden a (posibles) errores de carga. Por lo tanto, esos dos valores particulares se reemplazaron con nulos.

\paragraph{Treatment of missing values}

%De las 3 opciones para tratar las variables con valores nulos, se  descarta la de reemplazo con un valor especial único, ya que sería poco probable que éste fuese representativo para la persona, especialmente para las variables de laboratorio. Sí podría intentarse el uso de alguna técnica específica para determinar el valor más probable (p.e. regresor), pero su implementación queda por fuera del alcance de este trabajo.

%Se debe optar entre: (1) eliminar las variables que tienen nulos; o (2) eliminar los registros que tienen variables con valores nulos. La opción 1 permite mantener el tamaño muestral a costa de reducir la cantidad de variables de entrada para los modelos. En sentido opuesto, la opción 2 permite mantener la cantidad de características a costo de reducir el tamaño muestral.  En este caso, se decidió seleccionar la opción 2 únicamente para la variable \texttt{le\_diag\_\_\_familiar}, que cuenta con sólo 3 valores nulos. Por otro lado, la opción 1 se aplicó a las variables \texttt{edad}, \texttt{imc} y \texttt{circ\_de\_cintura}, considerando que hay una  variable cualitativa asociada que permite conocer el rango de cada valor. Para el resto de las variables, se discutirá en la Sección~\ref{sec:seg-prop}.

This task has two options: (1) removing variables with missing values; or (2) deleting records with null variables. The first option allows us to keep sample size at the cost of reducing the number of input variables for the models. The second option allows us to keep the number of features at the cost of reducing sample size. It was decided to select Option 2 only for the \texttt{family\_diagnosis} variable, which has only 3 null values. Option 1 was applied to the variables \texttt{age}, \texttt{bmi} and \texttt{waist\_circumference}, considering that there is an associated qualitative variable for each one of them that allows us to know its approximate value. All other variables will be discussed in Section~\ref{sec:seg-prop}.

\paragraph{Aggregation of class variable}

Table~\ref{tab:my-table2} shows that the distribution of classes is not balanced (\texttt{ogtt\_result}). To minimize the impact of this issue, a new class variable is created that splits subjects at risk of having PD or T2D (records with a value of “PD” or “T2D”) and those who are risk-free (records with a value of “Normal”). As a result, the variable is balanced in terms of the number of occurrences of each value, simplifying subsequent analyses as it now becomes a binary classification problem. However, caution should be used when analyzing the results, especially concerning the prediction of T2D, as it is the least common.

%De la Tabla~\ref{tab:my-table2} se puede notar que la distribución de clases no está balanceada (variable \texttt{resultado\_ptog}). Para minimizar el impacto de esta cuestión, se procede a crear una nueva variable de clase que divida entre personas sin riesgo de tener PDM o DM (registros con valor ``Normal'') y las que sí lo tienen (registros con valor ``PDM'' o ``DM''). Como resultado, la variable queda balanceada en cuanto a ocurrencias de cada valor y, a la vez, se simplifica el análisis posterior al pasar a ser ahora un problema de clasificación binaria. Sin embargo, se deberá ser cuidadoso al momento de analizar los resultados, especialmente con lo que se pueda decir sobre predicción de DM, por ser la de menor ocurrencia. 

\paragraph{Variable removal}

 %\texttt{resultado\_ptog} fue descartada por el agrupamiento realizado en la sección anterior, mientras que \texttt{glucemia\_pprandial} fue excluida, ya que contar con ese valor implicaría que la persona tuviera que hacerse una PTOG, careciendo de sentido el uso de los modelos propuestos.

 \texttt{ogtt\_result} was discarded in the previous section, and  \texttt{postprandial\_glycemia} was excluded because any subject with that variable would require an OGTT and, in that case, using the proposed models would be meaningless.

\subsubsection{Correlation Analyses}

%La Fig.~\ref{fig:corrmat} muestra la matriz de correlación obtenida sobre el \textit{dataset} inicial disponible. Desde el punto de vista clínico, tiene sentido que exista una correlación lineal débil entre el rango de la circunferencia de cintura y el rango IMC. También resulta razonable que existan correlaciones débiles entre los rangos de edad, IMC y circunferencia de cintura y sus valores asociados en las variables de \texttt{edad}, \texttt{imc} y \texttt{circ\_\_\_de\_cintura}. Por otra parte, el \texttt{colesterol\_total} se calcula a partir del \texttt{colesterol\_ldl}~\cite{aha_colesterol_2020}, lo que explica su correlación fuerte. Adicionalmente, la relación entre \texttt{glucemia\_basal}, \texttt{glucemia\_pprandial}, \texttt{resultado\_ptog} y \texttt{clase} tiene sentido, ya que las dos primeras determinan el valor de la tercera, la cual a su vez se agrupa para generar la cuarta de ellas.

\begin{figure}[tb]
         \centering
         \includegraphics[width=\textwidth]{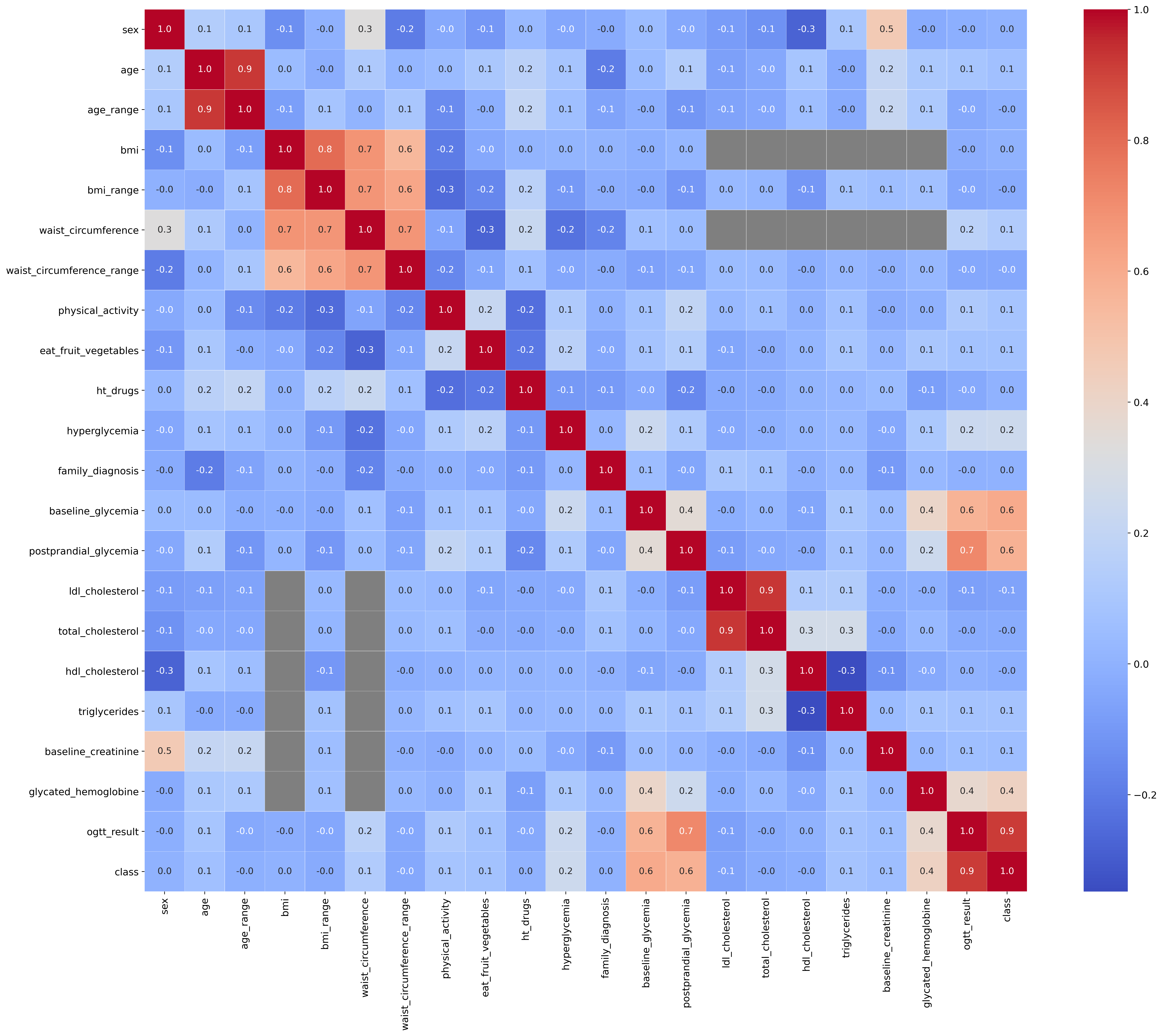}
         \caption{Correlation matrix for the initial dataset}
         \label{fig:corrmat}
     \end{figure}

Fig.~\ref{fig:corrmat} shows the correlation matrix obtained for the initial dataset. Clinically, it makes sense that there is a weak linear correlation between waist circumference range and BMI range. The weak correlation between age, BMI, and waist circumference ranges and their associated values for \texttt{age}, \texttt{bmi}, and \texttt{waist\_circumference} is also expected. On the other hand, \texttt{total\_cholesterol} is calculated from \texttt{ldl\_cholesterol}~\cite{aha_colesterol_2020}, which explains the strong correlation between these two variables. Additionally, the relationship between \texttt{baseline\_glycemia}, \texttt{postprandial\_glycemia}, \texttt{ogtt\_result} and, \texttt{class} is also to be expected, since the first two determine the value of the third, which in turn is grouped to generate the fourth.

\subsection{Proposed Segmentations}
\label{sec:seg-prop}

%Ante el porcentaje alto de nulidad de las variables de laboratorio, se planteó la posibilidad de generar varios datasets a partir del original, considerando distintos criterios para el tratamiento de registros nulos:

Given the high percentage of missing values in lab variables, the possibility of generating several datasets from the original by applying different criteria for the treatment of null records was considered:

\begin{comment}

\begin{itemize}
    \item \texttt{Dataset Clínica+Laboratorio (DCL-bin)}. Conjunto de datos al cual se realizó una eliminación de registros completos, derivando en 16 variables con 503 ejemplos (229 $\Rightarrow$ Sin riesgo, 274 $\Rightarrow$ Con riesgo). Este dataset mantiene todos las variables disponible (datos clínicos y de laboratorio) a costa de perder cantidad de registros. %Las tablas (Tabla~\ref{exploraciondcl} y Tabla~\ref{exploraciondclcontinuas}) resumen los features contenidos en este conjunto de datos.
    \item \texttt{Dataset Clínica+Glucemia basal (DCG-bin)}. Conjunto de datos que mantiene la información clínica disponible y la única variable de laboratorio que no posee valores nulos (\texttt{glucemia\_basal}); el resto de las ellas fueron eliminadas. De esta forma, este dataset cuenta con 10 variables y 1233 ejemplos. En contraposición con \texttt{DCL} aquí se preserva la cantidad de registros frente al valor aportado por el resto de los variables de laboratorio. %Las tablas (Tabla~\ref{exploraciondcg} y Tabla~\ref{exploraciondcgcontinuas}) resumen los features contenidos en este conjunto de datos.
\end{itemize}

\end{comment}
\begin{itemize}
    \item \texttt{Clinical and Lab Dataset (CLD-bin)}. All complete records were eliminated, resulting in 16 variables with 503 examples (229 $\Rightarrow$  No risk, 274 $\Rightarrow$  At risk). This dataset keeps all available variables (clinical data and lab data) at the cost of losing many records.
    
    \item \texttt{Clinical and baseline Glycemia Dataset (CGD-bin)}. This dataset keeps all available clinical information and the only lab variable that does not have null values (\texttt{baseline\_glycemia}); the rest of the variables were removed. Thus, this dataset has 10 variables and 1233 examples. In contrast to \texttt{CLD}, \texttt{CGD} preserves the number of records as opposed to lab variables value.

     \item \texttt{Clinical Dataset (CD-bin)}. This dataset keeps all available clinical information and removes all lab variables. Thus, this dataset includes 9 variables and 1233 examples, preserving the number of records. By discarding all lab features, simpler and cost-free models can be trained.

\end{itemize}

%Las versiones binarias de los conjuntos de datos son \texttt{DCL-bin} y \texttt{DCG-bin} respectivamente, los cuales traducen el problema a uno de clasificación binaria.

\section{Experimental Results}
\label{sec:results}

\subsection{Experimental Design}

%Se llevó a cabo un proceso de exploración, preprocesamiento, entrenamiento y evaluación de diversos algoritmos utilizando \textit{scikit-learn}. En particular, se consideraron Logistic Regression (LR), Decision Tree (DT) con  \texttt{max\_depth = 5} y \texttt{class\_weight = balanced}, k-Nearest Neighbor (kNN) con \texttt{n\_neighbors = 7} y Random Forest (RF) con \texttt{max\_depth = 2}. Asimismo, se utilizó TensorFlow para la construcción de modelos de Artificial Neural Networks (ANN) con \texttt{epochs = 60} y \texttt{batch\_size = 16}. En particular, se construyó una ANN de una capa oculta con 100 neuronas y función de activación \texttt{ReLU}, con una regularización L2 de $0.1$ para controlar el sobreajuste. Se utilizó un optimizador \texttt{Adamax} con una tasa de aprendizaje de $0.001$. %Como función de pérdida se utiliza la entropía cruzada binaria, adecuada para problemas de clasificación binaria.

An exploration, preprocessing, training, and evaluation process of various algorithms was carried out using \textit{scikit-learn}. In particular, Logistic Regression (LR), Decision Tree (DT) with \texttt{max\_depth} = 5 and \texttt{class\_weight = balanced}, k-Nearest Neighbor (kNN) with \texttt{n\_neighbors = 7}, and Random Forest (RF) with \texttt{max\_depth = 2} were considered. Likewise, TensorFlow was used to build Artificial Neural Networks (ANN) models with \texttt{epochs = 60} and \texttt{batch\_size = 16}. Specifically, a single hidden layer ANN was built, with 100 neurons and the \texttt{ReLU} activation function, and L2 regularization = 0.1 to control overfitting. An \texttt{Adamax} optimizer was used with a learning rate of 0.001.

%Para la manipulación, análisis y visualización de los datos, se han utilizado Pandas, NumPy y Matplotlib.
\begin{comment}

Siguiendo lineamientos habituales del área, el conjunto de datos fue particionado en dos: entrenamiento (70\%) y evaluación (30\%). Para reducir posibles sesgos, se aplicó la técnica de validación
cruzada con muestreo estratificado aleatorio (\textit{StratifiedShuffleSplit, n\_splits=50}~\cite{sklearn\_crossvalidation}). 
 %Además, se reservó un 30\% del \textit{dataset} para el conjunto de evaluación. 
 Debido a que las variables de entrada poseen magnitudes diferentes, se aplicó una normalización \textit{min-max} para todos los modelos excepto para la ANN donde se utilizó una normalización estándar. % Para esta última, además se especificó \texttt{epochs = 60} y \texttt{batch\_size =16}.
 
\end{comment}
 
 Following common guidelines in the area, the dataset was partitioned into two: training (70\%) and evaluation (30\%). To reduce possible biases, the cross-validation technique with stratified random sampling (\texttt{StratifiedShuffleSplit, n\_splits=50}~\cite{sklearn_crossvalidation}) was applied. Since input variables are expressed in different magnitudes, a \textit{min-max} normalization was applied for all models except for the ANN, where a standard normalization was used.

%Todas las experimentaciones se realizaron sobre una computadora local utilizando \textit{Jupyter Notebook}~\cite{jupyter\_notebook}. El hardware disponible fue una CPU Intel Core i7 2.6 GHz Quad-Core, 16 GB de memoria RAM y sistema operativo macOS.

All experiments were carried out on a local computer using \textit{Jupyter Notebook}~\cite{jupyter_notebook}. The hardware used was an Intel Core i7 2.6 GHz Quad-Core CPU, 16 GB of RAM, and macOS operating system.

\subsection{Models and Results for \texttt{CLD-bin}}

%La Tabla~\ref{tab:res-dcl-bin} muestra las métricas de rendimiento para los diferentes modelos aplicados al dataset  \texttt{DCL-bin} considerando como clase positiva a ``Con riesgo''. De los 5 modelos, se puede observar que hay 3 opciones que obtienen valores de \textit{accuracy} superiores al 90\%, lo que significa que aproximadamente 9 de cada 10 de los registros evaluados del total, fueron clasificados correctamente (los desvíos son menores al 2\%). De estas 3 opciones, RF es el que obtiene la mejor \textit{accuracy}, seguido de DT y finalmente ANN. Mientras que el rendimiento de LR se encuentra cercano a los anteriores con 86\% de \textit{accuracy}, el correspondiente a kNN es pobre al alcanzar sólo 72\%.

Table~\ref{tab:res-dcl-bin} shows the performance metrics for the different models applied to the \texttt{CLD-bin} dataset, considering “At risk” as a positive class. As it can be seen, 3 of the 5 models yield accuracy values greater than 90\%, which means that approximately 9 out of 10 of the total records evaluated were classified correctly (deviations are less than 2\%). Of these 3 options, RF is the one that yields the best accuracy, followed by DT, and finally ANN. While the performance of LR is close to the previous ones with 86\% accuracy, that of kNN is poor, reaching only 72\%.

%En cuanto a \textit{precision}, los valores se condicen con los de \textit{accuracy}. RF, DT y ANN presentan  valores superiores al 90\%   para la clase de interés (``Con riesgo'') lo que significa que, con cualquiera de las opciones, a más de 9 personas a las que se les dice que tiene riesgo, realmente lo tiene.  En el mismo sentido, los 3 modelos mencionados valores de \textit{recall} cercanos al 91\%, lo que significa que sólo 1 de cada 10 personas que tiene riesgo, no es identificada por ellos.

As regards precision, the values are consistent with those of accuracy. RF, DT and ANN yield values above 90\% for the class of interest (“At risk”), which means that, with any of the options, more than 9 in 10 people are correctly told they are at risk. Similarly, these 3 models have recall values close to 91\%, which means that only 1 in 10 people who are at risk is not identified.

%Por último, al momento de seleccionar un modelo determinado,  puede ser interesante examinar los valores de AUC además de la \textit{accuracy}. En este caso, se puede notar que RL desplaza a ANN como uno de los 3 que obtiene los valores más altos de AUC, en comparación a los de mejor \textit{accuracy}.

Finally, when selecting a specific model, AUC values can be of interest, in addition to accuracy. In this case, it can be seen that RL displaces ANN as one of the 3 models with the highest AUC values, compared to those with the best accuracy.

\begin{table}[tb!]
\centering
\caption{Results (evaluation) for the models applied to \texttt{CLD-bin} (positive class = “At risk”)}
\label{tab:res-dcl-bin}
%\resizebox{\columnwidth}{!}{%
\begin{tabular}
{p{.12\columnwidth}p{.18\columnwidth}p{.18\columnwidth}p{.18\columnwidth}p{.18\columnwidth}p{.08\columnwidth}}
\toprule
\textbf{Model} &
  \textbf{Accuracy} &
  \textbf{Precision} &
  \textbf{Recall} &
  \textbf{F-score} &
  \textbf{AUC} \\ \midrule
RF  & 94.58 $\pm$ 1.51 & 98.87 $\pm$ 1.3 & 91.1 $\pm$ 2.93 & 94.79 $\pm$ 1.53 & 0.95 \\
DT  & 93.42 $\pm$ 1.59 & 96.02 $\pm$ 2.02 & 91.73 $\pm$ 2.65 & 93.79 $\pm$ 1.53 & 0.94 \\
ANN & 91.13 $\pm$ 1.98 & 92.29 $\pm$ 2.68 & 91.39 $\pm$ 2.77 & 91.79 $\pm$ 1.84 & 0.91 \\
LR  & 85.56 $\pm$ 3.21 & 90.5 $\pm$ 3.11  & 82.1 $\pm$ 4.81  & 86.01 $\pm$ 3.3 & 0.93 \\
kNN & 71.62 $\pm$ 3.82 & 75.53 $\pm$ 3.87 & 70.83 $\pm$ 5.75 & 72.97 $\pm$ 4 & 0.72 \\ 
\bottomrule
\end{tabular}
%}
\end{table}

\subsection{Models and Results for \texttt{CGD-bin} }

%La Tabla~\ref{tab:res-dcg-bin} muestra las métricas de rendimiento para los diferentes modelos aplicados al dataset  \texttt{DCG-bin} considerando como clase positiva a ``Con riesgo''. Al igual que con \texttt{DCL-bin}, RF, DT y ANN obtienen muy buenos valores de \textit{accuracy}, siendo superiores al 90\% (los desvíos son menores al 2\%). kNN vuelve a presentar un rendimiento pobre con 70\% de \textit{accuracy}, mientras que LR se encuentra entre los 3 anteriores y kNN, aunque con un rendimiento más bajo que con el dataset anterior.

Table~\ref{tab:res-dcg-bin} shows the performance metrics for the different models applied to the \texttt{CGD-bin} dataset, considering “At risk” as a positive class. As with \texttt{CLD-bin}, RF, DT, and ANN yielded very good accuracy values, greater than 90\% (deviations below 2\%). kNN once again has a poor performance with 70\% accuracy, while LR is between the previous 3 and kNN, although its performance is lower than with the previous dataset.

%En cuanto a \textit{precision}, RF sobreajusta al obtener 100\% para la clase ``Con riesgo''. Por su parte, DT y ANN presentan  valores superiores al 90\% para la misma clase, estando LR y kNN bastante debajo de ese valor. Al analizar los valores de \textit{recall}, tanto RF como DT y ANN son los que consiguen los mejores valores,estando cercanos al 90\% para la clase de interés.

In terms of precision, RF overfits and obtains 100\% for the “At Risk” class. DT and ANN yield values above 90\% for the same class, with LR and kNN being well below this value. When analyzing recall values, RF, DT and, ANN are the models that achieve the best values, being close to 90\% for the class of interest.

%Por último, y a diferencia de \texttt{DCL-bin}, los 3 que obtienen los valores más altos de AUC coinciden con los de mejor \textit{accuracy}.

Finally, and unlike \texttt{CLD-bin}, the 3 models that obtain the highest AUC values are the same as the ones that achieve the best accuracy.

% \usepackage{booktabs}
\begin{table}[tb!]
\centering
\caption{Results (evaluation) for the models applied to \texttt{CGD-bin} (positive class = “At risk”)}
\label{tab:res-dcg-bin}
\begin{tabular}
{p{.12\columnwidth}p{.18\columnwidth}p{.18\columnwidth}p{.18\columnwidth}p{.18\columnwidth}p{.08\columnwidth}}
\toprule
\textbf{Model} & \textbf{Accuracy} & \textbf{Precision} & \textbf{Recall} & \textbf{F-score} & \textbf{AUC} \\ \midrule
RF  & 93.23 $\pm$ 1.12 & 100 $\pm$ 0   & 86.38 $\pm$ 2.25 & 92.68 $\pm$ 1.3 & 0.93 \\
DT  & 91.88 $\pm$ 1.67 & 96.68 $\pm$ 2.74 & 86.72 $\pm$ 2.2 & 91.39 $\pm$ 1.73 & 0.92 \\
ANN & 91.04 $\pm$ 1.63 & 94.45 $\pm$ 2.47 & 87.17 $\pm$ 2.31 & 90.64 $\pm$ 1.71 & 0.91 \\
LR  & 75.66 $\pm$ 4.37 & 80.96 $\pm$ 4.44 & 66.7 $\pm$ 6.55  & 73.05 $\pm$  5.3 & 0.83 \\
kNN & 69.97 $\pm$ 2.01 & 70.63 $\pm$ 2.32 & 67.92 $\pm$ 3.29 & 69.2 $\pm$ 2.23 & 0.7  \\ 
\bottomrule
\end{tabular}
\end{table}

\subsection{Models and Results for \texttt{CD-bin} }

Table~\ref{tab:res-dc-bin} shows the performance metrics for the different models applied to the \texttt{CD-bin} dataset, considering “At risk” as a positive class. Unlike the previous cases, the models present poor performances, reaching accuracy values from 54\% to 57\%. Moreover, the values presented for precision, recall, and F-score are along the same lines or worse. This means that the prediction is not much better than just flipping a coin.

% \usepackage{booktabs}
\begin{table}[tb!]
\centering
\caption{Results (evaluation) for the models applied to \texttt{CD-bin} (positive class = “At risk”)}
\label{tab:res-dc-bin}
\begin{tabular}
{p{.12\columnwidth}p{.2\columnwidth}p{.2\columnwidth}p{.2\columnwidth}p{.2\columnwidth}}
\toprule
\textbf{Model} & \textbf{Accuracy} & \textbf{Precision} & \textbf{Recall} & \textbf{F-score}  \\ \midrule
RF & 57.33 $\pm$ 1.93 & 63.77 $\pm$ 4.13 & 33.42 $\pm$ 4.84 & 43.61 $\pm$ 4.06  \\
DT  & 54.98 $\pm$ 2.47 & 55.3 $\pm$ 3.07 & 51.16 $\pm$ 9.03 & 52.67 $\pm$ 5.01 \\
ANN  & 57.55 $\pm$ 2.07 & 58.61 $\pm$ 3.06   & 51.05 $\pm$ 6.43 & 54.27 $\pm$ 3.71 \\
LR  & 57.16 $\pm$ 2.32 & 60.88 $\pm$ 4.27 & 39.92 $\pm$ 5.25  & 47.94 $\pm$  3.81 \\
kNN & 54.55 $\pm$ 2.45 & 54.45 $\pm$ 2.53 & 53.32 $\pm$ 4.79 & 53.77 $\pm$ 3.03  \\ 
\bottomrule
\end{tabular}
\end{table}

\subsection{Discussion}

Some of the models developed were able to achieve very good performances for both datasets. In particular, RF, DT, and ANN demonstrated great classification power, with high values in the considered metrics. In this sense, it should be noted that the proposed models are not intended to replace OGTT for the diagnosis of T2D and PD. As early detection is difficult for these diseases, the models are aimed at identifying individuals in the Argentine population who have a high probability of being affected and are unaware of their condition. To confirm the diagnosis, individuals identified as positive will eventually have to take an OGTT. The models would help identify those who should take the test and would make up for the lack of this type of tool.

It can be observed that there are no significant differences between the best accuracy and F-score values achieved for \texttt{CLD-bin} and \texttt{CGD-bin} datasets. Even though it is not (entirely) correct to compare results from models trained with different datasets, this issue could have an impact on the cost of putting the models into practice, considering that obtaining lab variables is neither free nor easy. To throw some light on this, a larger number of records without null values would be required.

Concerning the above, removing all lab features would lead to a simpler, cost-free model, that could be carried out at any time. Unfortunately, the performance results show that it is not feasible with the current \texttt{CD-bin} dataset. In that sense, it can be appreciated the great influence of the variable \texttt{baseline\_glycemia} on the success of the prediction, probably due to the limited size of the dataset.  Again, more records could compensate for the absence of lab variables.

Finally, it should also be taken into account that grouping PD and T2D into a single class favored balancing and simplified the problem by turning it into a binary classification. The cost of this decision is not being able to differentiate between PD and T2D. However, from a medical point of view, this would not be so relevant since, at the end of the day, the goal is to identify individuals at risk, regardless of which of the two conditions affects them.

\section{Conclusions and Future Work}
\label{sec:conc}

%Considerando que DM y PDM son enfermedades de difícil detección, en este trabajo se desarrollaron y evaluaron modelos predictivos específicos para la población argentina a partir de la base de datos del PPDBA. En primer lugar, fue necesario realizar un cuidadoso preprocesamiento de la base de datos, lo que derivó en la generación de dos datasets particulares (\texttt{DCL-bin} y \texttt{DCG-bin}) considerando el compromiso entre cantidad de variables y de registros disponibles. Luego, se aplicaron 5 modelos de clasificación diferentes a cada uno de ellos. Los resultados obtenidos muestran que  algunos de los modelos propuestos obtuvieron muy buenos rendimientos para ambos datasets. En particular, RF, DT y ANN demostraron gran poder de clasificación, con  valores altos en las métricas consideradas. Debido a limitaciones propias de la base de datos, no es posible afirmar que los resultados sean concluyentes, aunque  sí resultan promisorios. Considerando la vacancia de herramientas de este tipo para la población argentina, este trabajo representa el primer paso hacia modelos más sofisticados.

Considering that T2D and PD are difficult to detect, specific predictive models were developed and evaluated for the Argentine population based on the PPDBA database. Firstly, the database was carefully preprocessed, which led to the generation of three datasets (\texttt{CLD-bin}, \texttt{CGD-bin}, and \texttt{CD-bin}) with different approaches about the compromise between the number of variables and available records. Then, 5 different classification models were applied to each of them. The results obtained show that some of the models proposed were able to achieve very good performances for the first two datasets. In particular, RF, DT and ANN demonstrated great classification power, with high values for the metrics under consideration. In the opposite sense, removing all lab features led to poor performance results. Due to database limitations, the results are not conclusive, but they are promising. Given the lack of this type of tool in Argentina, this work represents the first step towards more sophisticated models.

\begin{comment}
Entre las líneas de trabajo futuro se encuentran:
\begin{itemize}
    \item Conseguir más registros de la base de datos para mejorar su calidad y al mismo tiempo aumentar su representatividad, para luego replicar el estudio realizado. 
    \item Evaluar el rendimiento de modelos generados a partir de una nueva segmentación que sólo considere datos clínicos. Un modelo de estas características sería más sencillo, sin costo y factible de realizar en cualquier momento, aunque probablemente de menor rendimiento.
    \item Considerar el desarrollo de modelos de clasificación multiclase para separar los casos de DM y PDM.
\end{itemize}
\end{comment}

Some of the future lines of work are:
\begin{itemize}
    \item Obtaining more database records to improve its quality and representativeness, and then replicating the study carried out.
 %   \item  Evaluating the performance of models generated using a new segmentation that only considers clinical data. A model with these characteristics would be simpler, cost-free, and could be carried out at any time; however, its performance would likely be lower.
    \item Considering the usage of additional supervised learning methods (like Support Vector Machines) and the hyper-parameter tuning of all trained models.
    \item 	Analyzing the development of multiclass classification models to separate T2D and PD cases.
\end{itemize}

\small
\bigskip \noindent\textbf{Funding.} This study was partially supported by PICT-2020-SERIEA-00901.


\bibliographystyle{splncs04}
\bibliography{references}

\end{document}